\documentclass[10pt,twocolumn,letterpaper]{article}

\PassOptionsToPackage{table,dvipsnames}{xcolor}
\usepackage[pagenumbers]{iccv}
\usepackage{multirow}

\definecolor{iccvblue}{rgb}{0.21,0.49,0.74}
\usepackage[breaklinks,colorlinks,allcolors=iccvblue]{hyperref}

\title{GrabVG: Graph-Attentive Binding for Visual Grounding in UAV Imagery}

\author{
Chaowei Wang\textsuperscript{1,*}\quad
Yan Di\textsuperscript{2,*}\quad
Jingjun Sun\textsuperscript{1}\quad
Baozhe Liu\textsuperscript{3}\\
Jiaxu Tian\textsuperscript{1}\quad
Yuheng Li\textsuperscript{1}\quad
Guangqian Guo\textsuperscript{1}\quad
Shan Gao\textsuperscript{1,\ensuremath{\dagger}}\\[3pt]
\textsuperscript{1}Northwestern Polytechnical University\quad
\textsuperscript{2}Harbin Institute of Technology\\
\textsuperscript{3}The Hong Kong Polytechnic University\\[2pt]
{\tt\small chaowei\_wang@mail.nwpu.edu.cn \quad
diyan@hit.edu.cn \quad
gaoshan@nwpu.edu.cn}\\[3pt]
{\small \textsuperscript{*}Equal contribution.\quad
\textsuperscript{\ensuremath{\dagger}}Corresponding author.}
}

\hypersetup{
  pdftitle={GrabVG: Graph-Attentive Binding for Visual Grounding in UAV Imagery},
  pdfauthor={Chaowei Wang, Yan Di, Jingjun Sun, Baozhe Liu, Jiaxu Tian, Yuheng Li, Guangqian Guo, Shan Gao}
}

\begin{document}

\maketitle

\begin{abstract}
Visual grounding in Unmanned Aerial Vehicle (UAV) imagery aims to localize a target object in complex bird's-eye-view scenes according to a natural language description. 
However, the abundance of small, densely distributed, and visually similar objects creates high visual redundancy, while repetitive local configurations give rise to strong topological ambiguity.
Existing approaches mainly focus on visual--language feature alignment or dense contextual interaction, yet they struggle to distinguish subtle inter-instance differences and effectively exploit spatial topological structures, leading to inaccurate grounding in highly crowded scenarios.
To address these challenges, we propose \textbf{GrabVG}, a novel visual grounding framework inspired by human visual search. 
GrabVG explicitly decomposes grounding into two sequential stages: \textit{preattentive hypothesis search} and \textit{graph-attentive feature binding}.
Specifically, we first generate a compact set of reliable object hypotheses through distillation-guided proposal induction and text-aware hypothesis filtering, substantially reducing background distractions and semantic mismatches. 
These hypotheses are then organized into a sparse graph, where language-guided intra-instance visual cues and inter-instance topological relationships are jointly bound and propagated via graph attention, enabling efficient spatial reasoning and accurate target localization.
Extensive experiments on AerialVG and AerialSense show that GrabVG achieves a favorable accuracy--speed trade-off, reaching 67.31\% and 80.34\% Acc@0.5 and outperforming the corresponding baselines by 10.55 and 8.76 percentage points, respectively.

\end{abstract}

\begin{figure}[t!]
    \centering
    \includegraphics[width=\columnwidth]{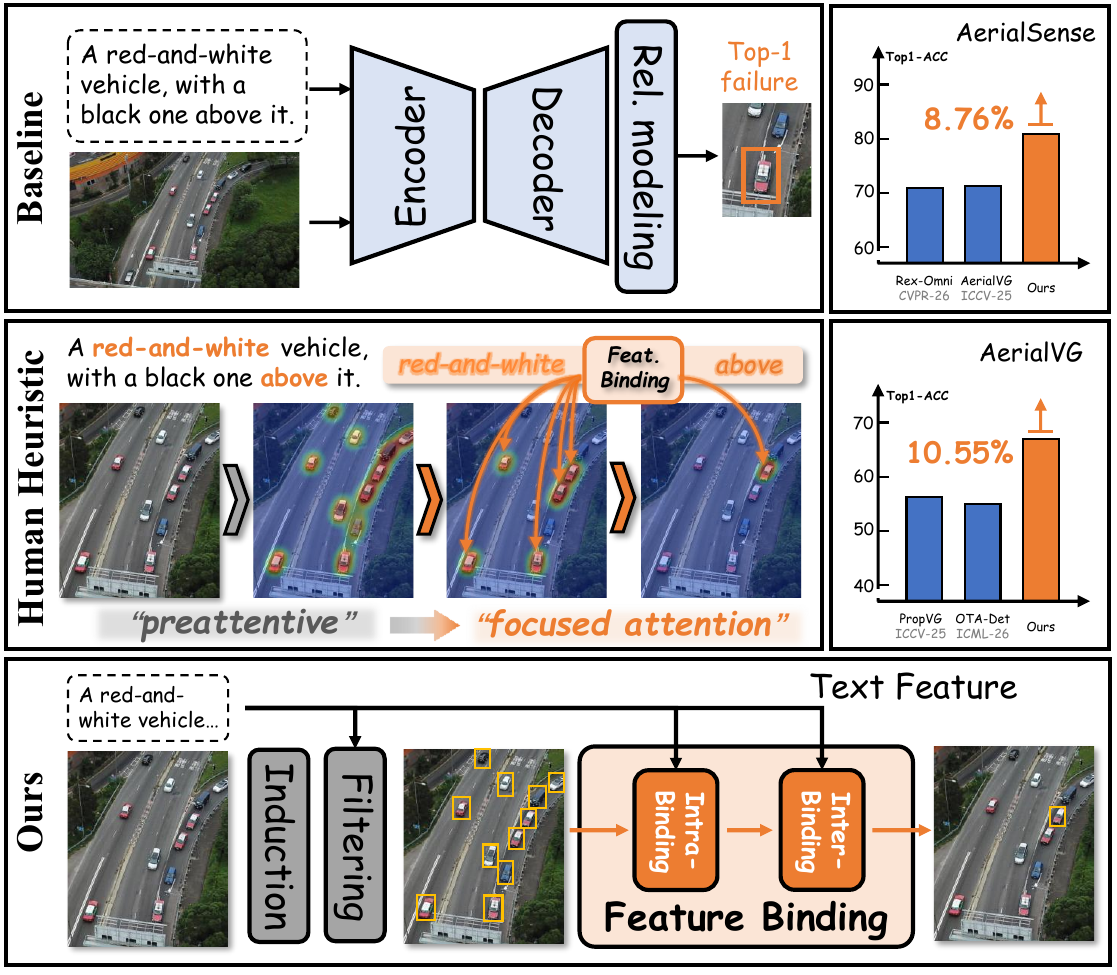} 
    \caption{
        Comparison of visual grounding paradigms.
        \textbf{Baseline~\cite{aerialvg}:} Existing methods rely on global decoding or dense relational modeling, often leaving visually similar hypotheses insufficiently distinguished.
        \textbf{Human Visual Search~\cite{wolfe1994,treisman1980}:} Preattentive guidance first narrows the search, followed by focused binding of discriminative evidence.
        \textbf{Ours:} GrabVG follows this progressive process through hypothesis induction, hypothesis filtering, and graph-attentive feature binding.
    }
    \label{fig:intro_comparison}
\end{figure}

\section{Introduction}

Visual grounding in Unmanned Aerial Vehicle (UAV) imagery aims to localize a target object in a bird's-eye-view scene according to a natural-language expression~\cite{aerialvg,aerialsense,aerialmid,trackbench}.
Compared with conventional ground-level imagery, UAV scenes typically contain numerous small and densely distributed objects, many of which exhibit highly similar appearances~\cite{dota,visionmeetsdrones,denseuav,aerialvg}.
The limited visual distinctions among these instances create \emph{high visual redundancy}, making appearance cues alone insufficient for distinguishing the referred object from nearby distractors~\cite{aerialvg}.
Moreover, the elevated viewpoint weakens perspective and depth cues while bringing a larger number of objects into view simultaneously. 
As a result, objects are often compressed into dense and repetitive two-dimensional arrangements, causing different target hypotheses to share similar local neighborhoods and relative spatial configurations. 
Such structurally similar neighborhoods make topological cues alone insufficient to uniquely identify the referent, giving rise to \emph{strong topological ambiguity}.

Existing visual grounding approaches address these challenges primarily through stronger vision--language alignment and contextual relation modeling.
Vision--language alignment methods enhance expression-conditioned visual representations~\cite{transvg,otadet}, while relation-based approaches incorporate spatial configurations among instances~\cite{aerialvg}.
However, cross-modal alignment is often performed across numerous intermediate queries, which may insufficiently encode subtle candidate-level differences.
Meanwhile, dense relation modeling captures extensive pairwise interactions, potentially allowing distant or weakly relevant relations to interfere with informative local context.
Consequently, existing approaches may still struggle to obtain discriminative candidate representations and reliable contextual evidence in crowded UAV scenes.
These limitations arise from both the modeled evidence and the organization of the grounding process.

Human visual search offers a useful principle for organizing this process.
Guided Search suggests that preattentive feature signals bias visual attention toward a restricted set of likely target locations~\cite{wolfe1994}, while Feature Integration Theory posits that focused attention is required to bind distributed visual features into coherent object representations~\cite{treisman1980}.
Inspired by this progressive process, we formulate UAV visual grounding as \emph{preattentive hypothesis search} followed by \emph{graph-attentive feature binding}: the former constrains the search space to plausible target hypotheses, whereas the latter associates each surviving hypothesis with discriminative evidence.

As illustrated in Fig.~\ref{fig:intro_comparison}, we instantiate this formulation as \textbf{GrabVG}, which separates candidate-space construction from evidence-based referent selection.
During \textit{Preattentive Hypothesis Search}, Distillation-Guided Proposal Induction transfers
expression-aligned object proposals from an external teacher pipeline, while Text-Aware Hypothesis Filtering filters out likely background queries and retains foreground candidates aligned with the referring expression.
During \textit{Graph-Attentive Feature Binding}, the remaining hypotheses are organized as a sparse geometric graph.
Intra-Instance Appearance Binding supplements each graph node with language-guided visual details sampled from the corresponding region and its immediate surroundings, whereas Inter-Instance Topological Binding exchanges edge-conditioned spatial information among neighboring nodes.
This separation allows GrabVG to preserve subtle instance-level appearance cues while limiting relational reasoning to geometrically meaningful neighborhoods, thereby reducing interference from dense all-pairs interactions.

Extensive experiments on AerialVG and AerialSense demonstrate the effectiveness and efficiency of GrabVG.
It achieves $67.31\%$ Acc@0.5 on AerialVG and $80.34\%$ Acc@0.5 on AerialSense, outperforming the corresponding baselines by $10.55$ and $8.76$ percentage points, respectively, while maintaining competitive inference speed.
Our main contributions are summarized as follows:

\begin{itemize}
    \item We reformulate crowded UAV visual grounding as a search-before-binding problem, separating the construction of plausible object candidates from the comparison of appearance and relational evidence. This formulation provides an explicit alternative to directly decoding the referent from dense visual tokens or unrestricted object interactions.

    \item We develop two complementary mechanisms for this formulation. Distillation-guided proposal induction and text-aware filtering establish a well-constrained, expression-aligned candidate space, while language-guided adaptive sampling and edge-aware graph attention enhance node appearance and model local geometric relations, respectively.

    \item Comprehensive experiments on AerialVG and AerialSense demonstrate the effectiveness and efficiency of GrabVG. Ablation studies further examine the contributions of proposal supervision, candidate filtering, adaptive appearance sampling, graph topology, and different pseudo-annotation sources.
\end{itemize}

\section{Related Work}

\paragraph{Visual Grounding.}
Visual grounding localizes an image region according to a natural-language expression.
Early proposal-based methods rank detected regions using appearance, location, and contextual cues~\cite{cmn,mattnet}, while one-stage and transformer-based approaches directly regress the referred region from cross-modal representations~\cite{fanet,rccf,transvg,mdetr,seqtr}.
Subsequent studies improve visual--language correspondence through language-guided feature refinement~\cite{qrnet,vglaw}, dynamic visual sampling~\cite{dynamic}, language-irrelevant token removal~\cite{scanformer}, and decoupled multimodal fusion~\cite{simvg}.
Grounded pre-training further unifies language-conditioned localization with open-vocabulary detection~\cite{glip,groundingdino}, while PropVG revisits proposal-driven grounding~\cite{propvg} and recent work explores progressive refinement for small-object REC~\cite{recsmall}. 
However, most of these methods either operate on dense multimodal features or rank generic region proposals, without explicitly separating candidate screening from instance-level relational reasoning.

\paragraph{UAV and Remote-Sensing Visual Grounding.}
Remote-sensing visual grounding is challenged by scale variation, cluttered backgrounds, high-resolution imagery, and small targets.
Existing studies improve cross-modal alignment~\cite{rsvg,geochat,otadet}, relational reasoning~\cite{aerialvg,dvgbench}, and semantic--geometric modeling~\cite{geovis,recs4r}.
ProVG further adopts a progressive \emph{survey--locate--verify} strategy that sequentially injects global, relational, and attribute cues into dense visual features~\cite{provg_rs}.
In contrast, GrabVG first screens explicit object candidates and then reasons over their local visual attributes and geometric neighborhoods.

\paragraph{Reasoning-Centric Remote-Sensing Grounding.}
Recent methods such as RSGround-R1 and Geo-R1 improve grounding through supervised reasoning traces and reinforcement fine-tuning~\cite{rsgroundr1,geor1}.
These approaches offer explicit textual rationales and improved low-shot reasoning, but require additional reasoning supervision or rollout-based optimization and incur autoregressive inference overhead.
GrabVG instead performs fixed-depth differentiable reasoning over sparse candidate graphs, offering directly inspectable node--edge interactions without generating textual reasoning trajectories.

\paragraph{Relation Modeling for Visual Grounding.}
Relational context is commonly used to distinguish visually similar candidates.
Early methods compare target proposals with surrounding objects or decompose expressions into subject, location, and relation components~\cite{cmn,mattnet}.
Later graph-based methods employ language-guided node and edge attention~\cite{neighbourhood}, cross-modal relation graphs~\cite{cmrin}, dynamic multi-step reasoning~\cite{dga}, or aligned language and visual scene graphs~\cite{sgmn}.
However, these relations are typically modeled over generic proposals or broad semantic structures.
AerialVG explicitly captures positional relations among aerial objects~\cite{aerialvg}, but dense interaction may introduce context from distant or weakly relevant instances.
GrabVG instead applies language-conditioned graph attention to filtered candidates connected by local geometric proximity, thereby limiting message passing from distant or weakly relevant instances.

\begin{figure*}[t]
  \centering
  \includegraphics[width=\linewidth]{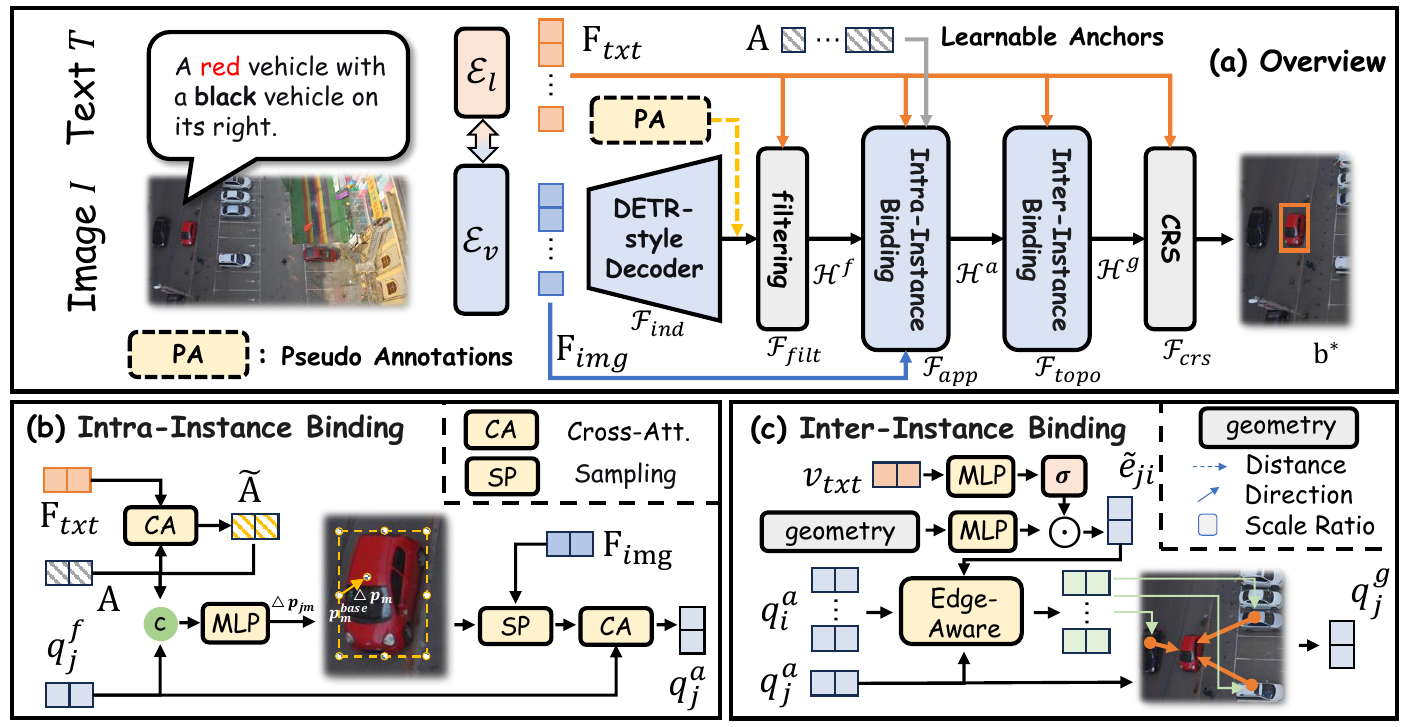} 
    \caption{\textbf{Overall architecture of GrabVG.}
    (a) Given a UAV image and a referring expression, Distillation-Guided Proposal Induction first produces an initial hypothesis set $\mathcal{H}^{0}$. Text-Aware Hypothesis Filtering then removes background queries and retains an expression-aligned candidate subset. The resulting hypotheses $\mathcal{H}^{f}$ are processed by Intra-Instance Appearance Binding and Inter-Instance Topological Binding, followed by CRS for final target selection.
    (b) Intra-Instance Appearance Binding samples language-guided visual features from each hypothesis region and its immediate surroundings to obtain appearance-aware representations $\mathbf{Q}^{a}$.
    (c) Inter-Instance Topological Binding exchanges relational information over a sparse geometric graph using language-conditioned edge features, producing graph-aware representations $\mathbf{Q}^{g}$.}
    \label{fig:framework_overview}
\end{figure*}

\section{Method}
\label{sec:method}

\subsection{Overview}

Given a UAV image $\mathbf{I}\in\mathbb{R}^{H\times W\times 3}$ and a referring expression $\mathbf{T}=\{w_l\}_{l=1}^{L}$, visual grounding aims to localize the unique instance described by $\mathbf{T}$ with a normalized bounding box $\mathbf{b}^{*}\in[0,1]^4$. As illustrated in Fig.~\ref{fig:framework_overview}, 
GrabVG resolves the referent through explicit candidate construction and graph-based comparison rather than direct decoding from an unconstrained set of visual tokens.

The vision and language encoders first produce multi-scale image features and token-level text features:
\begin{equation}
\begin{aligned}
    \mathbf{F}_{img}
    &=\{\mathbf{F}_{img}^{s}\}_{s=1}^{S}
    =\mathcal{E}_{v}(\mathbf{I}),\\
    \mathbf{F}_{txt}
    &=\mathcal{E}_{l}(\mathbf{T}).
\end{aligned}
\label{eq:grab_features}
\end{equation}
A DETR-style proposal generator $\mathcal{F}_{ind}$~\cite{detr,deformabledetr} first maps the multi-scale visual features $\mathbf{F}_{img}$ to $N$ initial hypotheses, $\mathcal{H}^{0}=\mathcal{F}_{ind}(\mathbf{F}_{img})=\{(\mathbf{q}_{i}^{0},\mathbf{b}_{i}^{0})\}_{i=1}^{N}$, where each hypothesis consists of a decoder representation and its normalized bounding box. GrabVG then progressively filters and binds these hypotheses through four consecutive states:
\begin{equation}
\begin{aligned}
    \mathcal{H}^{0}
    &\xrightarrow{\ \mathcal{F}_{filt}\ }
    \mathcal{H}^{f}
    \xrightarrow{\ \mathcal{F}_{app}\ }
    \mathcal{H}^{a}
    \xrightarrow{\ \mathcal{F}_{topo}\ }
    \mathcal{H}^{g}.
\end{aligned}
\label{eq:grab_pipeline}
\end{equation}
Here, Text-Aware Hypothesis Filtering $\mathcal{F}_{filt}$ scores the initial hypotheses using $\mathbf{F}_{txt}$ and retains $K$ foreground candidates, forming $\mathcal{H}^{f}$.
Intra-Instance Appearance Binding $\mathcal{F}_{app}$ updates their representations with sampled local image features to obtain $\mathcal{H}^{a}$.
Inter-Instance Topological Binding $\mathcal{F}_{topo}$ subsequently exchanges language-conditioned messages over neighboring candidates, yielding $\mathcal{H}^{g}$.
The bounding boxes remain unchanged throughout the two binding modules.

Let $\mathcal{H}^{g}=\{(\mathbf{q}_{j}^{g},\mathbf{b}_{j})\}_{j=1}^{K}$. We adopt the Contrastive-based Refer Scoring (CRS) module from PropVG~\cite{propvg} for final referent selection. CRS evaluates the compatibility between each graph-aware hypothesis and the referring expression to produce a referring score, while leaving the associated box unchanged:
\begin{equation}
\begin{aligned}
    s_{j}^{ref}
    &=\mathcal{F}_{crs}
    (\mathbf{q}_{j}^{g},\mathbf{F}_{txt}),\\
    \mathbf{b}^{*}
    &=\mathbf{b}_{j^{*}},
    \qquad
    j^{*}=\arg\max_{j}s_{j}^{ref}.
\end{aligned}
\label{eq:grab_output}
\end{equation}

\subsection{Preattentive Hypothesis Search}
Grounding annotations identify only the referred instance and provide little supervision for representing other plausible foreground objects.
Consequently, proposal queries trained solely with the final referring objective may receive insufficient supervision to form well-constrained object hypotheses in crowded scenes.
Preattentive Hypothesis Search addresses this issue in two steps: proposal induction transfers expression-aligned supervision from offline pseudo annotations, and text-aware filtering removes background queries before the more costly binding operations.

\paragraph{Distillation-Guided Proposal Induction.}
To supplement this sparse supervision, 
we follow the principle of teacher--student knowledge transfer~\cite{distillation} and employ an external teacher pipeline $\mathcal{F}_{tea}$~\cite{aerialvg,groundingdino} to generate an expression-aligned pseudo-annotation set offline: 
\begin{equation}
\begin{aligned}
    \mathcal{Y}_{tea}
    &=
    \left(
    \{\overline{\mathbf{b}}_{r}\}_{r=1}^{R},
    r^{*}
    \right)\\
    &=
    \mathcal{F}_{tea}(\mathbf{I},\mathbf{T}),
\end{aligned}
\label{eq:teacher_labels}
\end{equation}
where $\overline{\mathbf{b}}_{r}$ denotes a pseudo foreground box and $r^{*}$ identifies the pseudo annotation corresponding to the referred instance. These annotations are generated only for the training split and are not used during validation or inference. To guarantee positive referent supervision during training, the ground-truth referent box is appended when no retained teacher prediction satisfies the predefined matching criterion.

Using $\mathbf{F}_{img}$ as the visual memory, the proposal generator $\mathcal{F}_{ind}$ employs $N$ learnable queries and stacked query self-attention and multi-scale deformable cross-attention layers to produce the initial hypothesis set $\mathcal{H}^{0}$ defined in the Overview. A foreground classification head and a box regression head are attached to each decoder query. Their predictions are assigned to $\{\overline{\mathbf{b}}_{r}\}_{r=1}^{R}$ through one-to-one bipartite matching and optimized using the standard DETR classification, $\ell_{1}$ regression, and generalized IoU losses. 
This supervision encourages the decoder queries to cover multiple expression-aligned foreground hypotheses with accurate boxes, rather than concentrating only on the annotated referent.

\paragraph{Text-Aware Hypothesis Filtering.}
Although distillation improves proposal coverage, $\mathcal{H}^{0}$ may still contain background regions and hypotheses that are weakly aligned with the referring expression. We therefore employ a lightweight text-aware filter to perform coarse candidate screening before the more expensive binding operations. 
Its goal is to retain foreground hypotheses aligned with the referring expression, while fine-grained attribute and relational disambiguation is delegated to the subsequent binding and selection modules.

Let $\mathbf{Q}^{0}=[\mathbf{q}_{1}^{0},\ldots,\mathbf{q}_{N}^{0}]$. A single cross-attention block~\cite{transformer} injects linguistic context into the proposal features for relevance estimation:
\begin{equation}
\begin{aligned}
    \mathbf{U}
    &=\operatorname{LN}\!\left(
    \mathbf{Q}^{0}+
    \operatorname{MHA}(\mathbf{Q}^{0},\mathbf{F}_{txt},\mathbf{F}_{txt})
    \right),\\
    \overline{\mathbf{Q}}
    &=\operatorname{LN}\!\left(
    \mathbf{U}+\operatorname{FFN}(\mathbf{U})
    \right),\qquad
    \mathbf{s}^{f}=\sigma(\mathbf{W}_{f}\overline{\mathbf{Q}}).
\end{aligned}
\label{eq:filter_features}
\end{equation}
Here, $\operatorname{MHA}(\mathbf{Q},\mathbf{K},\mathbf{V})$ denotes a multi-head attention block. The top-$K$ scoring hypotheses are retained:
\begin{equation}
\begin{aligned}
    \Omega&=\operatorname{TopK}(\mathbf{s}^{f},K)
    =\{\omega_{1},\ldots,\omega_{K}\},\\
    \mathcal{H}^{f}
    &=\{(\mathbf{q}_{\omega_j}^{0},\mathbf{b}_{\omega_j}^{0})\}_{j=1}^{K}
    \triangleq\{(\mathbf{q}_{j}^{f},\mathbf{b}_{j})\}_{j=1}^{K}.
\end{aligned}
\label{eq:filter_selection}
\end{equation}
The auxiliary features $\overline{\mathbf{Q}}$ are used only to estimate filtering scores, while the selected original decoder features $\{\mathbf{q}_{\omega_j}^{0}\}_{j=1}^{K}$ are forwarded to the binding stage. This separates language-guided hypothesis selection from subsequent feature binding, with the filtering scores supervised by $\mathcal{L}_{filt}$.

\subsection{Graph-Attentive Feature Binding}
Graph-Attentive Feature Binding updates the graph through node-level visual sampling and edge-conditioned message passing.
Intra-Instance Appearance Binding enriches each node using features sampled around its associated box, whereas Inter-Instance Topological Binding exchanges spatial information among geometrically neighboring candidates.

\paragraph{Intra-Instance Appearance Binding.}
Within the hypothesis graph, each node initially contains a decoder feature $\mathbf{q}_{j}^{f}$ and its associated box $\mathbf{b}_{j}=(\mathbf{c}_{j},\boldsymbol{\rho}_{j})$, where $\mathbf{c}_{j}$ and $\boldsymbol{\rho}_{j}$ denote the normalized box center and size, respectively. To supplement the decoder feature with candidate-specific visual evidence, we associate each hypothesis with $M$ box-relative anchors $\{(\mathbf{a}_{m},\mathbf{p}_{m}^{base})\}_{m=1}^{M}$, where $\mathbf{a}_{m}$ is a learnable query embedding and $\mathbf{p}_{m}^{base}$ denotes its predefined base location relative to the hypothesis box. These query embeddings are first conditioned on the referring expression, after which candidate-specific offsets are jointly predicted from the hypothesis representation and the resulting language-aware anchor embeddings:

\begin{equation}
\begin{aligned}
\mathbf{A}^{t}
&=
\operatorname{MHA}
\left(
\mathbf{A},
\mathbf{F}_{txt},
\mathbf{F}_{txt}
\right),\\
\Delta\mathbf{p}_{jm}
&=\operatorname{MLP}_{off}
\left(
[\mathbf{q}_{j}^{f};
 \mathbf{a}_{m};
 \mathbf{a}_{m}^{t}]
\right),\\
\mathbf{p}_{jm}
&=\operatorname{clip}_{[0,1]^2}
\left(
\mathbf{c}_{j}
+
\left(
\mathbf{p}_{m}^{base}
+\Delta\mathbf{p}_{jm}
\right)
\odot\boldsymbol{\rho}_{j}
\right).
\end{aligned}
\label{eq:appearance_sampling_compact}
\end{equation}
Here, $\mathbf{A}=[\mathbf{a}_{1},\ldots,\mathbf{a}_{M}]$ and $\mathbf{A}^{t}=[\mathbf{a}_{1}^{t},\ldots,\mathbf{a}_{M}^{t}]$ denote the original and language-aware anchor embeddings, respectively. The predicted locations therefore adapt to both the candidate representation and the linguistic context while remaining relative to the candidate box. We then sample multi-scale visual features at these locations and aggregate them to update the corresponding hypothesis:
\begin{equation}
\begin{aligned}
\mathbf{F}_{j}^{loc}&=\operatorname{Sample}\!\left(\mathbf{F}_{img},\{\mathbf{p}_{jm}\}_{m=1}^{M}\right),\\
\mathbf{q}_{j}^{a}
&=
\operatorname{MHA}
\left(
\mathbf{q}_{j}^{f},
[\mathbf{q}_{j}^{f};\mathbf{F}_{j}^{loc}],
[\mathbf{q}_{j}^{f};\mathbf{F}_{j}^{loc}]
\right).
\end{aligned}
\label{eq:appearance_update_compact}
\end{equation}
Here, $\operatorname{Sample}$ denotes differentiable bilinear sampling from all visual feature levels followed by cross-scale feature fusion. The resulting appearance-aware hypothesis set is $\mathcal{H}^{a}=\{(\mathbf{q}_{j}^{a},\mathbf{b}_{j})\}_{j=1}^{K}$. 

\paragraph{Inter-Instance Topological Binding.}
After appearance refinement, the hypothesis boxes remain unchanged while the node attributes are replaced by $\{\mathbf{q}_{j}^{a}\}_{j=1}^{K}$. To capture local spatial configurations without introducing distracting all-pairs interactions, we construct a symmetric $k$-nearest-neighbor graph from the detached normalized hypothesis centers and include a self-loop for every node:
\begin{equation}
\mathcal{N}(j)=\operatorname{SymKNN}_{k}(\mathbf{c}_{j})\cup\{j\}.
\label{eq:knn_graph}
\end{equation}
The self-loop allows each node to directly balance its own appearance evidence against messages from neighboring hypotheses during attention aggregation. The graph connectivity is shared across all graph-attention layers. For each directed edge $j\leftarrow i$, we construct a geometric descriptor $\mathbf{r}_{ji}$ from the relative center displacement, distance, direction, and log-scale difference between the two boxes.
A pooled sentence representation $\mathbf{v}_{txt}$ then modulates the projected geometric descriptor:
\begin{equation}
\widetilde{\mathbf{e}}_{ji}=\operatorname{MLP}_{geo}(\mathbf{r}_{ji})\odot\sigma\!\left(\operatorname{MLP}_{txt}(\mathbf{v}_{txt})\right).
\label{eq:edge_feature_compact}
\end{equation}
Starting from $\mathbf{z}_{j}^{(0)}=\mathbf{q}_{j}^{a}$, we perform edge-aware graph attention over the local neighborhood. For clarity, the attention-head index is omitted:
\begin{equation}
\begin{aligned}
a_{ji}^{\ell}
&=
\frac{1}{\sqrt{d}}
\left(
\mathbf{W}_{q}^{\ell}
\mathbf{z}_{j}^{(\ell-1)}
\right)^{\top}
\left(
\mathbf{W}_{k}^{\ell}
\mathbf{z}_{i}^{(\ell-1)}
+
\mathbf{W}_{ek}^{\ell}
\widetilde{\mathbf{e}}_{ji}
\right)\\
&\quad+
\left(
\mathbf{w}_{eb}^{\ell}
\right)^{\top}
\widetilde{\mathbf{e}}_{ji},\\
\alpha_{ji}^{\ell}
&=
\operatorname{Softmax}_{i\in\mathcal{N}(j)}
\left(
a_{ji}^{\ell}
\right),\\
\mathbf{m}_{j}^{\ell}
&=
\sum_{i\in\mathcal{N}(j)}
\alpha_{ji}^{\ell}
\left(
\mathbf{W}_{v}^{\ell}
\mathbf{z}_{i}^{(\ell-1)}
+
\mathbf{W}_{ev}^{\ell}
\widetilde{\mathbf{e}}_{ji}
\right),\\
\mathbf{z}_{j}^{(\ell)}
&=
\operatorname{GraphBlock}^{\ell}
\left(
\mathbf{z}_{j}^{(\ell-1)},
\mathbf{m}_{j}^{\ell}
\right).
\end{aligned}
\label{eq:graph_attention_compact}
\end{equation}

The edge-conditioned key and bias terms determine the attention allocated to neighboring hypotheses, while the edge-conditioned value injects relational information into the aggregated message $\mathbf{m}_{j}^{\ell}$. The graph block then combines this message with the previous node state through residual and feed-forward updates. After $L_g$ graph-attention layers, a learned sigmoid gate adaptively blends the graph-propagated state $\mathbf{z}_{j}^{(L_g)}$ with the appearance-aware representation $\mathbf{q}_{j}^{a}$ to produce $\mathbf{q}_{j}^{g}$. The resulting graph-aware hypothesis set is $\mathcal{H}^{g}=\{(\mathbf{q}_{j}^{g},\mathbf{b}_{j})\}_{j=1}^{K}$.

\subsection{Training Objective}

GrabVG is jointly optimized with proposal-induction, hypothesis-filtering, grounding, and auxiliary segmentation objectives:
\begin{equation}
\begin{aligned}
\mathcal{L}
=&\lambda_{det}\mathcal{L}_{det}
+\lambda_{filt}\mathcal{L}_{filt}\\
&+\lambda_{ref}\mathcal{L}_{ref}
+\lambda_{mask}\mathcal{L}_{mask}.
\end{aligned}
\label{eq:training_objective}
\end{equation}
The proposal-induction loss $\mathcal{L}_{det}$ follows DETR~\cite{detr} and combines foreground classification, $\ell_{1}$ box regression, and generalized IoU losses~\cite{giou}. The grounding loss $\mathcal{L}_{ref}$ and auxiliary segmentation loss $\mathcal{L}_{mask}$ follow PropVG~\cite{propvg}.

For hypothesis filtering, let $\mathcal{M}$ denote the bipartite matching between induced hypotheses and pseudo annotations. A hypothesis is treated as a positive filtering target if it is matched to any pseudo box. Let $i^{r}$ denote the hypothesis matched to the referred pseudo annotation and $\Omega$ the retained index set. We combine binary relevance supervision with a retention penalty that discourages pruning the referred hypothesis:
\begin{equation}
\begin{aligned}
y_i^{f}
&=
\mathbb{I}
\left[
\exists r,\,
(i,r)\in\mathcal{M}
\right],\\
\mathcal{L}_{filt}
&=
\lambda_{bce}
\operatorname{BCE}
(\mathbf{s}^{f},\mathbf{y}^{f})\\
&\quad+
\lambda_{keep}
\mathbb{I}[i^{r}\notin\Omega]
\left[-\log(s_{i^{r}}^{f}+\epsilon)\right].
\end{aligned}
\label{eq:filter_loss}
\end{equation}
The filtering objective performs coarse foreground screening, while final referent discrimination is supervised by $\mathcal{L}_{ref}$ over the retained hypotheses.

\section{Experiments}
\subsection{Setup}

\paragraph{Datasets.}
We evaluate GrabVG on \textbf{AerialVG}~\cite{aerialvg} and the visual-grounding subset of \textbf{AerialSense}~\cite{aerialsense}. AerialVG contains 5,000 high-resolution images, 50,000 referring expressions, and 103,000 bounding boxes, and we follow its official train--test split. AerialSense covers diverse resolutions, illumination conditions, and UAV scenes; as no official split is available, we use the first $75\%$ of samples for training and the remaining $25\%$ for testing.

\paragraph{Metrics.}
We report \textbf{Acc@0.5 (Top1)}, \textbf{Top-5 Acc@0.5 (Top5)}, and \textbf{mIoU}. Top1 and Top5 measure whether the top-1 prediction or any of the five highest-scoring boxes has IoU above $0.5$, respectively, while mIoU is the mean IoU of the top-1 predictions.

\paragraph{Baselines.}
Our comparisons cover general visual grounding methods~\cite{propvg}, UAV-specific visual grounding methods~\cite{aerialvg,otadet}, and zero-shot grounding methods~\cite{rexomni}.

\paragraph{Implementation Details.}
Following PropVG~\cite{propvg}, we adopt the BEiT-3 Base~\cite{beit3} backbone and a three-layer deformable proposal decoder. Images are resized with the aspect ratio preserved and padded to $586\times586$, and the maximum text length is $64$. We initialize $N=256$ hypotheses for AerialVG and $N=128$ for AerialSense, of which the top-scoring $50\%$ are retained. Appearance binding uses $48$ adaptive anchors, while topological binding employs four graph-attention layers over a symmetrized 6-NN graph. The models are trained using Adam for $36$ epochs on AerialVG and $45$ epochs on AerialSense, with learning rates of $1.5\times10^{-4}$ for the newly introduced modules and $1.5\times10^{-5}$ for the backbone. Unless otherwise stated, we use AerialVG-derived pseudo annotations for AerialVG and Grounding-DINO-derived pseudo annotations for AerialSense.
The filtering layer uses eight attention heads. Training uses a batch size of $3$ per GPU, gradient clipping at $0.15$, and two epochs of linear warm-up. The learning rate is subsequently reduced by a factor of $0.1$; for AerialVG, the decay is applied at epochs $24$ and $32$.

\subsection{Comparison with State of the Art}

\paragraph{Results on AerialVG.}
As shown in Table~\ref{tab:sota_aerialvg}, we compare the proposed GrabVG with recent methods, including both zero-shot multimodal approaches and fully supervised aerial-specific models. GrabVG achieves state-of-the-art performance on the AerialVG test set, with $67.31\%$ Top1, $89.43\%$ Top5, and $53.34\%$ mIoU. Compared with the strongest one-stage baseline, OTA-Det ($54.90\%$), and the dense-relation model AerialVG ($50.03\%$), GrabVG improves Top1 by $12.41$ and $17.28$ percentage points, respectively. Furthermore, GrabVG outperforms the hyperparameter-optimized two-stage baseline PropVG* by $10.55$ percentage points in Top1. These results suggest that expression-aligned proposal supervision, local visual sampling, and neighborhood-restricted graph reasoning provide complementary benefits in crowded UAV scenes.


\begin{table}[t]
\centering
\resizebox{\linewidth}{!}{
\begin{tabular}{l c c c}
\toprule
\textbf{Method} & \textbf{Top1 (\%)} & \textbf{Top5 (\%)} & \textbf{mIoU (\%)} \\
\midrule
\multicolumn{4}{l}{\textit{Zero-Shot Methods}} \\
ReX-Omni~\cite{rexomni} & 28.37 & 30.09 & - \\
\midrule
\multicolumn{4}{l}{\textit{Supervised Methods}} \\
TransVG~\cite{transvg} & 11.53 & 13.68 & - \\
D-MDETR~\cite{dynamic} & 19.87 & 29.87 & - \\
G-DINO~\cite{groundingdino} & 29.36 & 78.87 & - \\
AerialVG~\cite{aerialvg} & 50.03 & 87.00 & - \\
PropVG~\cite{propvg} & 49.40 & - & 49.38 \\
PropVG*~\cite{propvg} & 56.76 & 74.05 & 46.76 \\
OTA-Det~\cite{otadet} & 54.90 & - & - \\
\rowcolor{gray!10} \textbf{GrabVG (Ours)} & \textbf{67.31} & \textbf{89.43} & \textbf{53.34} \\
\bottomrule
\end{tabular}
}
\caption{\textbf{Comparison with state-of-the-art methods on the AerialVG test set.} ``*'' indicates results obtained with hyperparameter optimization. Best results are in \textbf{bold}. Top1 and Top5 denote Acc@0.5 and Top-5 Acc@0.5, respectively.}
\label{tab:sota_aerialvg}
\end{table}

\paragraph{Results on AerialSense.}
To evaluate the generalization capability of GrabVG across diverse UAV scenarios, we report results on AerialSense in Table~\ref{tab:sota_aerialsense}. GrabVG achieves the best performance, reaching $80.34\%$ Top1 and $72.00\%$ mIoU. It surpasses the zero-shot ReX-Omni and the supervised AerialVG baseline by $8.87$ and $8.76$ percentage points in Top1, respectively. These results demonstrate that the proposed candidate construction and graph-attentive binding framework generalizes effectively beyond AerialVG.

\begin{table}[t]
\centering
\resizebox{\linewidth}{!}{
\begin{tabular}{l c c c}
\toprule
\textbf{Method} & \textbf{Type} & \textbf{Top1 (\%)} & \textbf{mIoU (\%)} \\
\midrule
\multicolumn{4}{l}{\textit{Large Multimodal Models (Zero-Shot)}} \\
ReX-Omni~\cite{rexomni} & Zero-shot & 71.47 & - \\
\midrule
\multicolumn{4}{l}{\textit{Domain-Specific Models}} \\
AerialVG~\cite{aerialvg} & Supervised & 71.58 & - \\
PropVG*~\cite{propvg} & Supervised & 61.30 & 67.76 \\
\rowcolor{gray!10} \textbf{GrabVG (Ours)} & Supervised & \textbf{80.34} & \textbf{72.00} \\
\bottomrule
\end{tabular}
}
\caption{\textbf{Performance on the AerialSense benchmark.} We compare GrabVG with a recent zero-shot multimodal model and aerial-specific supervised baselines.}
\label{tab:sota_aerialsense}
\end{table}

\subsection{Comprehensive Ablation Study}
\paragraph{Component-wise Contribution.}
Table~\ref{tab:main_ablation} incrementally evaluates each component from the optimized PropVG baseline, which is trained on the original ground-truth annotations and achieves $56.76\%$ Top1.
Introducing \textbf{Distillation-Guided Proposal Induction} with expression-aligned pseudo annotations improves Top1 to $62.84\%$, indicating that denser foreground supervision produces a more reliable and better-covered hypothesis space.
Adding \textbf{Intra-Instance Appearance Binding} further improves Top1 by $2.75$ points to $65.59\%$, demonstrating the benefit of recovering candidate-specific local cues for distinguishing visually similar objects.
In comparison, \textbf{Inter-Instance Topological Binding} alone yields a $2.26$-point gain to $65.10\%$, showing that local geometric relations among neighboring hypotheses provide effective contextual evidence.
Combining both binding modules reaches $67.01\%$, confirming that fine-grained appearance cues and inter-instance relational information are complementary.
Finally, \textbf{Text-Aware Hypothesis Filtering} achieves the best Top1 of $67.31\%$ by suppressing weakly relevant hypotheses before feature binding.

\begin{table}[t]
\centering
\resizebox{\linewidth}{!}{
\begin{tabular}{c | c c c c | cc | cc}
\toprule
\multirow{2}{*}{\textbf{Baseline}} & \multicolumn{4}{c|}{\textbf{GrabVG Components}} & \multicolumn{2}{c|}{\textbf{Test}} & \multicolumn{2}{c}{\textbf{Val}} \\
\cmidrule{2-9}
& $\boldsymbol{\mathcal{Y}_{tea}}$ & $\boldsymbol{\mathcal{F}_{app}}$ & $\boldsymbol{\mathcal{F}_{topo}}$ & $\boldsymbol{\mathcal{F}_{filt}}$ & \textbf{Top1} & \textbf{Top5} & \textbf{Top1} & \textbf{Top5} \\
\midrule
\checkmark & & & & & 56.76 & 74.05 & 55.96 & 72.28 \\
\midrule
\checkmark & \checkmark & & & & 62.84 & 85.54 & 61.82 & 85.04 \\
\checkmark & \checkmark & \checkmark & & & 65.59 & 87.89 & 65.42 & 87.13 \\
\checkmark & \checkmark & & \checkmark &  & 65.10 & 88.02 & 64.15 & 87.86 \\
\checkmark & \checkmark & \checkmark & \checkmark & & 67.01 & \textbf{89.52} & 67.13 & \textbf{88.96} \\
\rowcolor{gray!10} \checkmark & \checkmark & \checkmark & \checkmark & \checkmark & \textbf{67.31} & 89.43 & \textbf{67.41} & 88.78 \\
\bottomrule
\end{tabular}
}
\caption{
\textbf{Component-wise Ablation on AerialVG.} The baseline is a hyperparameter-optimized PropVG model trained with the
original ground-truth annotations.
$\mathcal{Y}_{tea}$ denotes fixed expression-aligned pseudo annotations
generated offline by a trained AerialVG teacher;
$\mathcal{F}_{app}$, $\mathcal{F}_{topo}$, and $\mathcal{F}_{filt}$ denote
appearance binding, topological binding, and text-aware filtering,
respectively.
$\mathcal{F}_{topo}$ uses a symmetric $k$-NN graph.
}
\label{tab:main_ablation}
\end{table}

\subsection{Further Experimental Analysis}

\paragraph{Filtering Strategy.}
Table~\ref{tab:pipeline_design} examines where and how often hypotheses should be screened. All variants use the same AerialVG-derived pseudo annotations for proposal induction. Replacing the learned filter with non-maximum suppression (NMS) produces a similar Top1 result ($67.25\%$ versus $67.31\%$) and a higher Top5 result, but NMS does not exploit the referring expression. Applying the learned filter both before and after appearance binding reduces Top1 to $66.75\%$. Repeated pruning can therefore discard ambiguous but correct hypotheses before topological evidence becomes available. A single early text-aware filter provides the best Top1 while keeping the subsequent binding stages focused on a compact candidate set.

\begin{table}[t]
\centering
\resizebox{\linewidth}{!}{
\begin{tabular}{l c c}
\toprule
\textbf{Pipeline Strategy} & \textbf{Top1 (\%)} & \textbf{Top5 (\%)} \\
\midrule
$\mathcal{F}_{ind}+\mathcal{F}_{topo}$ (no appearance binding) & 65.10 & 88.02 \\
$\mathcal{F}_{ind}+\mathcal{F}_{app}+\mathcal{F}_{topo}+\mathrm{NMS}$ & 67.25 & \textbf{90.03} \\
$\mathcal{F}_{ind}+\mathcal{F}_{filt}+\mathcal{F}_{app}+\mathcal{F}_{filt}+\mathcal{F}_{topo}$ & 66.75 & 89.46 \\
\rowcolor{gray!10} $\mathcal{F}_{ind}+\mathcal{F}_{filt}+\mathcal{F}_{app}+\mathcal{F}_{topo}$ & \textbf{67.31} & 89.43 \\
\bottomrule
\end{tabular}
}
\caption{\textbf{Pipeline design choices on AerialVG.} Only the post-induction candidate-processing strategy is changed.}
\label{tab:pipeline_design}
\end{table}

\paragraph{Appearance-Binding Design.}
Table~\ref{tab:intra_binding} studies the spatial layout of the adaptive sampling anchors. These experiments isolate appearance binding by disabling filtering and topological binding. Single- and double-ring layouts achieve Top1 values between $64.36\%$ and $64.59\%$. The triple-ring layout with scales $[0.5,1.0,1.2]$ improves Top1 to $65.59\%$ and Top5 to $87.89\%$. Covering the hypothesis interior, boundary, and immediate surroundings therefore provides more discriminative evidence than concentrating all sampling points at one or two scales.

\begin{table}[t]
\centering
\resizebox{\linewidth}{!}{
\begin{tabular}{l c c c}
\toprule
\textbf{Anchor Topology} & \textbf{Scales} & \textbf{Top1 (\%)} & \textbf{Top5 (\%)} \\
\midrule
Single Ring & $[1.0]$ & 64.59 & 87.15 \\
Double Ring & $[0.5,1.0]$ & 64.57 & 87.51 \\
Double Ring & $[1.0,1.2]$ & 64.36 & 87.56 \\
\rowcolor{gray!10} Triple Ring (Ours) & $[0.5,1.0,1.2]$ & \textbf{65.59} & \textbf{87.89} \\
\bottomrule
\end{tabular}
}
\caption{\textbf{Ablation of Intra-Instance Appearance Binding on AerialVG.} All variants use proposal induction followed by appearance binding; filtering and topological binding are disabled.}
\label{tab:intra_binding}
\end{table}

\paragraph{Topology and Propagation Depth.}
Table~\ref{tab:inter_binding} compares dense interaction, Delaunay triangulation, and the default KNN graph while keeping the remaining pipeline fixed. Dense all-pairs interaction reaches only $63.98\%$ Top1, consistent with the hypothesis that distant or weakly relevant candidates introduce distracting context. Delaunay triangulation improves Top1 to $66.21\%$, whereas the 6-NN graph reaches $67.31\%$. Sparse local connectivity is therefore more effective than unrestricted interaction in crowded UAV scenes.
The depth comparison reveals a complementary trade-off. Increasing the graph from two to four layers improves Top1 by $2.97$ percentage points and produces the highest Top5 value. Six layers obtain the highest Top1 ($67.96\%$) but reduce Top5, while eight layers degrade both metrics. We use four layers because they provide the most balanced localization and candidate-recall performance without excessive feature propagation.

\begin{table}[t]
\centering
\resizebox{\linewidth}{!}{
\begin{tabular}{l c c c}
\toprule
\textbf{Relational Topology} & \textbf{Layers} & \textbf{Top1 (\%)} & \textbf{Top5 (\%)} \\
\midrule
\multicolumn{4}{l}{\textit{Topology Strategy}} \\
Dense (AerialVG-style) & 4 & 63.98 & 87.07 \\
\rowcolor{gray!10} KNN ($k=6$) & 4 & \textbf{67.31} & \textbf{89.43} \\
Delaunay Triangulation & 4 & 66.21 & 88.65 \\
\midrule
\multicolumn{4}{l}{\textit{Graph Depth with KNN ($k=6$)}} \\
KNN & 2 & 64.34 & 88.32 \\
\rowcolor{gray!10} KNN & 4 & 67.31 & \textbf{89.43} \\
KNN & 6 & \textbf{67.96} & 88.51 \\
KNN & 8 & 63.79 & 87.58 \\
\bottomrule
\end{tabular}
}
\caption{\textbf{Ablation of Inter-Instance Topological Binding on AerialVG.} The upper block changes the graph structure; the lower block changes only the depth of the default KNN graph.}
\label{tab:inter_binding}
\end{table}

\paragraph{Robustness to Hypothesis Priors.}
GrabVG can be trained with proposal supervision from different sources. Table~\ref{tab:prior_robustness} compares original training boxes, two open-vocabulary pseudo-annotation pipelines, AerialVG-derived pseudo annotations, and an oracle setting containing all ground-truth boxes. Relative to the corresponding source baseline, the complete refinement stack improves Top1 by $2.18$--$4.47$ percentage points for all non-oracle sources. With perfect oracle hypotheses, appearance and topological binding still yield a $4.61$-point gain even though filtering is omitted. These consistent improvements indicate that the binding modules are not tied to one teacher or proposal distribution; they remain useful under both imperfect and ideal candidate coverage.

\begin{table}[t]
\centering
\resizebox{\linewidth}{!}{
\begin{tabular}{l c c c}
\toprule
\textbf{Prior Source} & \textbf{Source Baseline} & \textbf{With GrabVG} & \textbf{Gain (pp)} \\
\midrule
Training GT Boxes & 56.76 & 58.94 & \textbf{+2.18} \\
LLM + Grounding-DINO~\cite{groundingdino} & 56.83 & 60.06 & \textbf{+3.23} \\
LLM + SAM3~\cite{sam3} & 56.97 & 60.36 & \textbf{+3.39} \\
AerialVG-Derived~\cite{aerialvg} & 62.84 & 67.31 & \textbf{+4.47} \\
\midrule
Oracle (All GT Boxes)$^{\ast}$ & 69.62 & 74.23 & \textbf{+4.61} \\
\bottomrule
\end{tabular}
}
\caption{\textbf{Robustness to hypothesis priors on the AerialVG test set.} All values are Top1 (Acc@0.5). LLM denotes Qwen3~\cite{qwen3}. $^{\ast}$The oracle setting supplies all ground-truth boxes and omits $\mathcal{F}_{filt}$.}
\label{tab:prior_robustness}
\end{table}

\paragraph{Inference Efficiency.}
Table~\ref{tab:efficiency} reports inference speed under the same evaluation setup.
AerialVG runs at $13.65$ FPS, whereas GrabVG reaches $28.51$ FPS and approaches the $32.00$ FPS of OTA-Det.
Without hypothesis filtering, $\mathcal{F}_{app}$ and $\mathcal{F}_{topo}$ run at $25.09$ FPS.
Introducing $\mathcal{F}_{filt}$ reduces the number of candidates processed by the two binding modules and increases the speed to $28.51$ FPS.
These results show that early candidate screening recovers part of the computational overhead introduced by graph-attentive binding.

\begin{table}[ht]
\centering
\resizebox{0.8\linewidth}{!}{
\begin{tabular}{l c}
\toprule
\textbf{Method} & \textbf{FPS (img/s)} \\
\midrule
\multicolumn{2}{l}{\textit{Existing Baselines}} \\
AerialVG~\cite{aerialvg} & 13.65 \\
OTA-Det~\cite{otadet} & 32.00 \\
\midrule
\multicolumn{2}{l}{\textit{GrabVG Variants}} \\
$\mathcal{F}_{ind}$ & 30.32 \\
\quad + $\mathcal{F}_{app}$ (Intra-binding) & 30.03 \\
\quad + $\mathcal{F}_{app}$ + $\mathcal{F}_{topo}$ (Inter-binding) & 25.09 \\
\quad + $\mathcal{F}_{filt}$ + $\mathcal{F}_{app}$ + $\mathcal{F}_{topo}$ (Full GrabVG) & 28.51 \\
\bottomrule
\end{tabular}
}
\caption{\textbf{Inference-speed comparison.}
FPS results for GrabVG, its component variants, and existing baselines under the same evaluation setup. All FPS results are measured on a single NVIDIA RTX 4090 GPU.}
\label{tab:efficiency}
\end{table}

\subsection{Qualitative Results}

\begin{figure}[htbp]
    \centering
    \includegraphics[width=\linewidth]{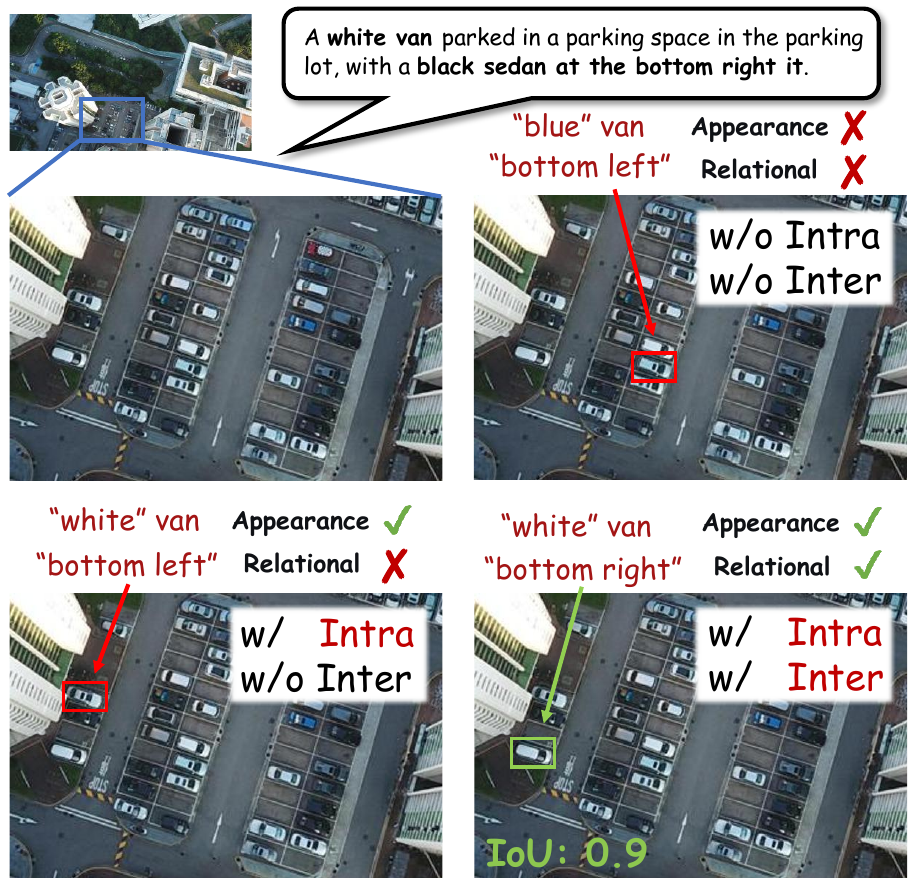} 
    \caption{\textbf{Qualitative ablation of progressive feature binding.} Appearance binding corrects the target attribute, while topological binding resolves the remaining spatial ambiguity.}
    \label{fig:ablation_qualitative}
\end{figure}

\paragraph{Progressive Disambiguation.}
Figure~\ref{fig:ablation_qualitative} illustrates how the two binding modules progressively resolve appearance and relational ambiguity. Without either module, the model selects a blue vehicle whose appearance and relative position are both inconsistent with the expression. After introducing \textbf{Intra-Instance Appearance Binding}, the prediction shifts to a white van, indicating that language-guided local visual sampling captures the referred appearance attribute; however, the selected instance still violates the specified ``bottom-right'' relation. With \textbf{Inter-Instance Topological Binding} further incorporated, the model propagates spatial evidence from neighboring hypotheses and identifies the white van whose surrounding configuration satisfies the expression.

\begin{figure}[htbp]
    \centering
    \includegraphics[width=\linewidth]{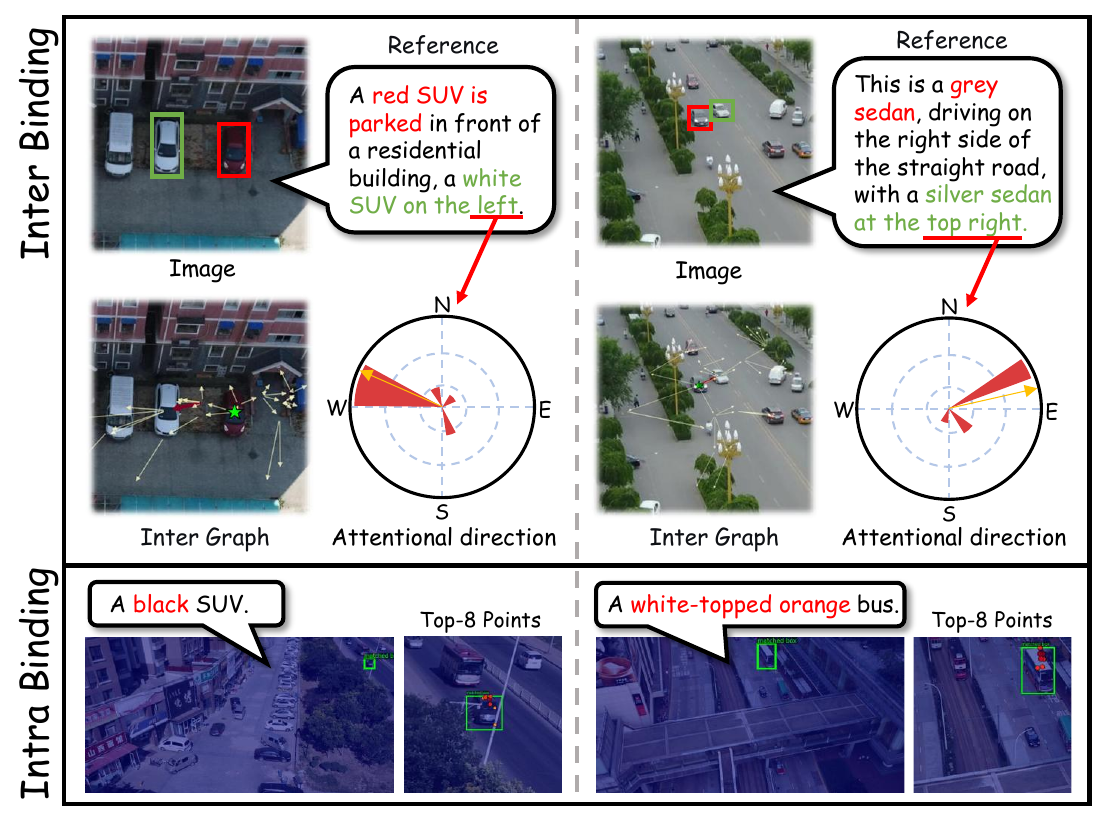}
    \caption{\textbf{Visualization of Graph-Attentive Feature Binding.} Top: sparse graph edges and directional attention distributions, where thick red arrows indicate the highest-attended edges. Bottom: top-$8$ adaptive sampling points highlighting appearance-relevant regions.}
    \label{fig:attention_visualization}
\end{figure}

\paragraph{Binding Mechanism Visualization.}
Figure~\ref{fig:attention_visualization} illustrates how GrabVG binds topological and appearance evidence. In the upper part, thin yellow arrows denote sparse graph edges, the thick red arrow marks the highest-attended edge, and the polar plots summarize the directions of highly weighted edges. For ``a white SUV on the left'' and ``a silver sedan at the top right,'' the dominant directions point west and northeast, respectively, showing that \textbf{Inter-Instance Topological Binding} emphasizes neighboring hypotheses consistent with the described spatial relations. In the lower part, the top-$8$ sampling points from \textbf{Intra-Instance Appearance Binding} concentrate on the referred objects and discriminative regions, such as the dark body of the black SUV and the white roof and orange body of the bus. Together, these visualizations show that the two modules capture expression-relevant relational directions and fine-grained local appearance evidence, respectively.

\section{Conclusion}

We presented GrabVG for visual grounding in crowded UAV imagery. The central idea is to separate candidate-space construction from evidence-based referent selection: expression-aligned pseudo-annotation supervision and text-aware filtering first establish a manageable candidate pool, after which adaptive visual sampling and sparse graph attention compare local appearance and geometric neighborhood information. This design limits the influence of background queries and distant candidate interactions while retaining cues needed to distinguish nearby, appearance-similar objects. GrabVG achieves $67.31\%$ Top1 on AerialVG and $80.34\%$ on AerialSense, exceeding the corresponding baselines by $10.55$ and $8.76$ percentage points. The ablation results further show that proposal supervision, appearance sampling, local graph propagation, and early filtering provide complementary improvements across the evaluated settings.
Future work will explore scene-adaptive retention and neighborhood settings.

{
    \small
    \bibliographystyle{ieeenat_fullname}
    \bibliography{main}

@inproceedings{propvg,
  title={Propvg: End-to-end proposal-driven visual grounding with multi-granularity discrimination},
  author={Dai, Ming and Cheng, Wenxuan and Zhuang, Jiedong and Liu, Jiang-jiang and Zhao, Hongshen and Feng, Zhenhua and Yang, Wankou},
  booktitle={Proceedings of the IEEE/CVF International Conference on Computer Vision},
  pages={7058--7068},
  year={2025}
}

@inproceedings{aerialvg,
  title={Aerialvg: A challenging benchmark for aerial visual grounding by exploring positional relations},
  author={Liu, Junli and Chen, Qizhi and Wang, Zhigang and Tang, Yiwen and Zhang, Yiting and Yan, Chi and Wang, Dong and Li, Xuelong and Zhao, Bin},
  booktitle={Proceedings of the IEEE/CVF International Conference on Computer Vision},
  pages={5177--5187},
  year={2025}
}

@article{aerialsense,
  title={Towards Accurate UAV Image Perception: Guiding Vision-Language Models with Stronger Task Prompts},
  author={Guo, Mingning and Wu, Mengwei and Li, Shaoxian and Li, Haifeng and Tao, Chao},
  journal={arXiv preprint arXiv:2512.07302},
  year={2025}
}

@article{dvgbench,
  title={DVGBench: Implicit-to-explicit visual grounding benchmark in UAV imagery with large vision--language models},
  author={Zhou, Yue and Chen, Jue and Zhang, Zilun and Huang, Penghui and Ding, Ran and Zou, Zhentao and Gao, PengFei and Wei, Yuchen and Li, Ke and Yang, Xue and others},
  journal={ISPRS Journal of Photogrammetry and Remote Sensing},
  volume={232},
  pages={831--847},
  year={2026},
  publisher={Elsevier}
}

@inproceedings{aerialmid,
  title={Aerialmind: Towards referring multi-object tracking in UAV scenarios},
  author={Chen, Chenglizhao and Liang, Shaofeng and Guan, Runwei and Sun, Xiaolou and Zhao, Haocheng and Jiang, Haiyun and Huang, Tao and Ding, Henghui and Han, Qing-Long},
  booktitle={Proceedings of the AAAI Conference on Artificial Intelligence},
  volume={40},
  number={4},
  pages={2805--2813},
  year={2026}
}

@article{trackbench,
  title={A benchmark for UAV-view natural language-guided tracking},
  author={Li, Hengyou and Liu, Xinyan and Li, Guorong},
  journal={Electronics},
  volume={13},
  number={9},
  pages={1706},
  year={2024},
  publisher={MDPI}
}

@article{otadet,
  title={Open-Text Aerial Detection: A Unified Framework For Aerial Visual Grounding And Detection},
  author={Wei, Guoting and Yuan, Xia and Zhou, Yang and Jing, Haizhao and Liu, Yu and Qi, Xianbiao and Zhao, Chunxia and Zhang, Haokui and Xiao, Rong},
  journal={arXiv preprint arXiv:2602.07827},
  year={2026}
}

@inproceedings{transvg,
  title={Transvg: End-to-end visual grounding with transformers},
  author={Deng, Jiajun and Yang, Zhengyuan and Chen, Tianlang and Zhou, Wengang and Li, Houqiang},
  booktitle={Proceedings of the IEEE/CVF international conference on computer vision},
  pages={1769--1779},
  year={2021}
}

@inproceedings{mattnet,
  title={Mattnet: Modular attention network for referring expression comprehension},
  author={Yu, Licheng and Lin, Zhe and Shen, Xiaohui and Yang, Jimei and Lu, Xin and Bansal, Mohit and Berg, Tamara L},
  booktitle={Proceedings of the IEEE conference on computer vision and pattern recognition},
  pages={1307--1315},
  year={2018}
}

@inproceedings{mdetr,
  title={Mdetr-modulated detection for end-to-end multi-modal understanding},
  author={Kamath, Aishwarya and Singh, Mannat and LeCun, Yann and Synnaeve, Gabriel and Misra, Ishan and Carion, Nicolas},
  booktitle={Proceedings of the IEEE/CVF international conference on computer vision},
  pages={1780--1790},
  year={2021}
}

@inproceedings{groundingdino,
  title={Grounding dino: Marrying dino with grounded pre-training for open-set object detection},
  author={Liu, Shilong and Zeng, Zhaoyang and Ren, Tianhe and Li, Feng and Zhang, Hao and Yang, Jie and Jiang, Qing and Li, Chunyuan and Yang, Jianwei and Su, Hang and others},
  booktitle={European conference on computer vision},
  pages={38--55},
  year={2024},
  organization={Springer}
}

@inproceedings{glip,
  title={Grounded language-image pre-training},
  author={Li, Liunian Harold and Zhang, Pengchuan and Zhang, Haotian and Yang, Jianwei and Li, Chunyuan and Zhong, Yiwu and Wang, Lijuan and Yuan, Lu and Zhang, Lei and Hwang, Jenq-Neng and others},
  booktitle={Proceedings of the IEEE/CVF conference on computer vision and pattern recognition},
  pages={10965--10975},
  year={2022}
}

@inproceedings{qrnet,
  title={Shifting more attention to visual backbone: Query-modulated refinement networks for end-to-end visual grounding},
  author={Ye, Jiabo and Tian, Junfeng and Yan, Ming and Yang, Xiaoshan and Wang, Xuwu and Zhang, Ji and He, Liang and Lin, Xin},
  booktitle={proceedings of the IEEE/CVF conference on computer vision and pattern recognition},
  pages={15502--15512},
  year={2022}
}

@inproceedings{beit3,
  title={Image as a foreign language: Beit pretraining for vision and vision-language tasks},
  author={Wang, Wenhui and Bao, Hangbo and Dong, Li and Bjorck, Johan and Peng, Zhiliang and Liu, Qiang and Aggarwal, Kriti and Mohammed, Owais Khan and Singhal, Saksham and Som, Subhojit and others},
  booktitle={Proceedings of the IEEE/CVF Conference on Computer Vision and Pattern Recognition},
  pages={19175--19186},
  year={2023}
}

@inproceedings{geochat,
  title={Geochat: Grounded large vision-language model for remote sensing},
  author={Kuckreja, Kartik and Danish, Muhammad Sohail and Naseer, Muzammal and Das, Abhijit and Khan, Salman and Khan, Fahad Shahbaz},
  booktitle={Proceedings of the IEEE/CVF conference on computer vision and pattern recognition},
  pages={27831--27840},
  year={2024}
}

@article{treisman1980,
  title={A feature-integration theory of attention},
  author={Treisman, Anne M and Gelade, Garry},
  journal={Cognitive psychology},
  volume={12},
  number={1},
  pages={97--136},
  year={1980},
  publisher={Elsevier}
}

@article{wolfe1994,
  title={Guided search 2.0 a revised model of visual search},
  author={Wolfe, Jeremy M},
  journal={Psychonomic bulletin \& review},
  volume={1},
  number={2},
  pages={202--238},
  year={1994},
  publisher={Springer}
}

@inproceedings{dota,
  title={DOTA: A large-scale dataset for object detection in aerial images},
  author={Xia, Gui-Song and Bai, Xiang and Ding, Jian and Zhu, Zhen and Belongie, Serge and Luo, Jiebo and Datcu, Mihai and Pelillo, Marcello and Zhang, Liangpei},
  booktitle={Proceedings of the IEEE conference on computer vision and pattern recognition},
  pages={3974--3983},
  year={2018}
}

@article{visionmeetsdrones,
  title={Vision meets drones: A challenge},
  author={Zhu, Pengfei and Wen, Longyin and Bian, Xiao and Ling, Haibin and Hu, Qinghua},
  journal={arXiv preprint arXiv:1804.07437},
  year={2018}
}

@inproceedings{denseuav,
  title={Dense and small object detection in UAV vision based on cascade network},
  author={Zhang, Xindi and Izquierdo, Ebroul and Chandramouli, Krishna},
  booktitle={Proceedings of the IEEE/CVF international conference on computer vision workshops},
  pages={0--0},
  year={2019}
}

@inproceedings{seqtr,
  title={Seqtr: A simple yet universal network for visual grounding},
  author={Zhu, Chaoyang and Zhou, Yiyi and Shen, Yunhang and Luo, Gen and Pan, Xingjia and Lin, Mingbao and Chen, Chao and Cao, Liujuan and Sun, Xiaoshuai and Ji, Rongrong},
  booktitle={European Conference on Computer Vision},
  pages={598--615},
  year={2022},
  organization={Springer}
}

@inproceedings{fanet,
  title={A fast and accurate one-stage approach to visual grounding},
  author={Yang, Zhengyuan and Gong, Boqing and Wang, Liwei and Huang, Wenbing and Yu, Dong and Luo, Jiebo},
  booktitle={Proceedings of the IEEE/CVF international conference on computer vision},
  pages={4683--4693},
  year={2019}
}

@inproceedings{dga,
  title={Dynamic graph attention for referring expression comprehension},
  author={Yang, Sibei and Li, Guanbin and Yu, Yizhou},
  booktitle={Proceedings of the IEEE/CVF international conference on computer vision},
  pages={4644--4653},
  year={2019}
}

@inproceedings{sgmn,
  title={Graph-structured referring expression reasoning in the wild},
  author={Yang, Sibei and Li, Guanbin and Yu, Yizhou},
  booktitle={Proceedings of the IEEE/CVF conference on computer vision and pattern recognition},
  pages={9952--9961},
  year={2020}
}

@article{dynamic,
  title={Dynamic MDETR: A Dynamic Multimodal Transformer Decoder for Visual Grounding},
  author={Shi, Fengyuan and Gao, Ruopeng and Huang, Weilin and Wang, Limin},
  journal={IEEE Transactions on Pattern Analysis and Machine Intelligence},
  volume={46},
  number={2},
  pages={1181--1198},
  year={2024},
  publisher={IEEE}
}

@inproceedings{rexomni,
  title={Detect anything via next point prediction},
  author={Jiang, Qing and Huo, Junan and Chen, Xingyu and Xiong, Yuda and Zeng, Zhaoyang and Chen, Yihao and Ren, Tianhe and Yu, Junzhi and Zhang, Lei},
  booktitle={Proceedings of the IEEE/CVF Conference on Computer Vision and Pattern Recognition},
  pages={25472--25483},
  year={2026}
}

@inproceedings{cmn,
  title={Modeling context in referring expressions},
  author={Yu, Licheng and Poirson, Patrick and Yang, Shan and Berg, Alexander C and Berg, Tamara L},
  booktitle={European conference on computer vision},
  pages={69--85},
  year={2016},
  organization={Springer}
}

@inproceedings{rccf,
  title={A real-time cross-modality correlation filtering method for referring expression comprehension},
  author={Liao, Yue and Liu, Si and Li, Guanbin and Wang, Fei and Chen, Yanjie and Qian, Chen and Li, Bo},
  booktitle={proceedings of the IEEE/CVF conference on computer vision and pattern recognition},
  pages={10880--10889},
  year={2020}
}

@inproceedings{vglaw,
  title={Language adaptive weight generation for multi-task visual grounding},
  author={Su, Wei and Miao, Peihan and Dou, Huanzhang and Wang, Gaoang and Qiao, Liang and Li, Zheyang and Li, Xi},
  booktitle={Proceedings of the IEEE/CVF conference on computer vision and pattern recognition},
  pages={10857--10866},
  year={2023}
}

@inproceedings{scanformer,
  title={Scanformer: Referring expression comprehension by iteratively scanning},
  author={Su, Wei and Miao, Peihan and Dou, Huanzhang and Li, Xi},
  booktitle={Proceedings of the IEEE/CVF conference on computer vision and pattern recognition},
  pages={13449--13458},
  year={2024}
}

@article{simvg,
  title={Simvg: A simple framework for visual grounding with decoupled multi-modal fusion},
  author={Dai, Ming and Yang, Lingfeng and Xu, Yihao and Feng, Zhenhua and Yang, Wankou},
  journal={Advances in neural information processing systems},
  volume={37},
  pages={121670--121698},
  year={2024}
}

@inproceedings{recsmall,
  title={Referring Expression Comprehension for Small Objects},
  author={Goto, Kanoko and Hirose, Takumi and Ukai, Mahiro and Kurita, Shuhei and Inoue, Nakamasa},
  booktitle={Proceedings of the IEEE/CVF International Conference on Computer Vision},
  pages={21231--21242},
  year={2025}
}

@article{rsvg,
  title={Rsvg: Exploring data and models for visual grounding on remote sensing data},
  author={Zhan, Yang and Xiong, Zhitong and Yuan, Yuan},
  journal={IEEE transactions on geoscience and remote sensing},
  volume={61},
  pages={1--13},
  year={2023},
  publisher={IEEE}
}

@article{rsgroundr1,
  title={RSGround-R1: Rethinking Remote Sensing Visual Grounding through Spatial Reasoning},
  author={Huang, Shiqi and He, Shuting and Wen, Bihan},
  journal={arXiv preprint arXiv:2601.21634},
  year={2026}
}

@inproceedings{geovis,
  title={Geovis: Geospatially rewarded visual search for remote sensing visual grounding},
  author={Zhang, Peirong and Zhang, Yidan and Xu, Luxiao and Lin, Jinliang and Guo, Zonghao and Wang, Fengxiang and Yang, Xue and Wei, Kaiwen and Wang, Lei},
  booktitle={Proceedings of the IEEE/CVF Conference on Computer Vision and Pattern Recognition},
  pages={14335--14345},
  year={2026}
}

@inproceedings{recs4r,
  title={RECS4R: Bridging Semantics and Geometry for Referring Remote Sensing Interpretation},
  author={Chai, Jinming and Li, Lingling and Jiao, Licheng and Lu, Xiaoqiang and Sun, Long and Liu, Xu and Ma, Wenping and Li, Weibin},
  booktitle={Proceedings of the IEEE/CVF Conference on Computer Vision and Pattern Recognition},
  pages={42213--42224},
  year={2026}
}

@inproceedings{neighbourhood,
  title={Neighbourhood watch: Referring expression comprehension via language-guided graph attention networks},
  author={Wang, Peng and Wu, Qi and Cao, Jiewei and Shen, Chunhua and Gao, Lianli and Hengel, Anton van den},
  booktitle={proceedings of the IEEE/CVF conference on computer vision and pattern recognition},
  pages={1960--1968},
  year={2019}
}

@inproceedings{cmrin,
  title={Cross-modal relationship inference for grounding referring expressions},
  author={Yang, Sibei and Li, Guanbin and Yu, Yizhou},
  booktitle={Proceedings of the IEEE/CVF conference on computer vision and pattern recognition},
  pages={4145--4154},
  year={2019}
}

@article{provg_rs,
  title={Provg: Progressive visual grounding via language decoupling for remote sensing imagery},
  author={Li, Ke and Wang, Ting and Wang, Di and Zhu, Yongshan and Zhang, Yiming and Lei, Tao and Wang, Quan},
  journal={arXiv preprint arXiv:2604.01893},
  year={2026}
}

@inproceedings{detr,
  title={End-to-end object detection with transformers},
  author={Carion, Nicolas and Massa, Francisco and Synnaeve, Gabriel and Usunier, Nicolas and Kirillov, Alexander and Zagoruyko, Sergey},
  booktitle={European conference on computer vision},
  pages={213--229},
  year={2020},
  organization={Springer}
}

@article{deformabledetr,
  title={Deformable detr: Deformable transformers for end-to-end object detection},
  author={Zhu, Xizhou and Su, Weijie and Lu, Lewei and Li, Bin and Wang, Xiaogang and Dai, Jifeng},
  journal={arXiv preprint arXiv:2010.04159},
  year={2020}
}

@article{distillation,
  title={Distilling the knowledge in a neural network},
  author={Hinton, Geoffrey and Vinyals, Oriol and Dean, Jeff},
  journal={arXiv preprint arXiv:1503.02531},
  year={2015}
}

@article{transformer,
  title={Attention is all you need},
  author={Vaswani, Ashish and Shazeer, Noam and Parmar, Niki and Uszkoreit, Jakob and Jones, Llion and Gomez, Aidan N and Kaiser, {\L}ukasz and Polosukhin, Illia},
  journal={Advances in neural information processing systems},
  volume={30},
  year={2017}
}

@inproceedings{giou,
  title={Generalized intersection over union: A metric and a loss for bounding box regression},
  author={Rezatofighi, Hamid and Tsoi, Nathan and Gwak, JunYoung and Sadeghian, Amir and Reid, Ian and Savarese, Silvio},
  booktitle={Proceedings of the IEEE/CVF conference on computer vision and pattern recognition},
  pages={658--666},
  year={2019}
}

@article{qwen3,
  title={Qwen3 technical report},
  author={Yang, An and Li, Anfeng and Yang, Baosong and Zhang, Beichen and Hui, Binyuan and Zheng, Bo and Yu, Bowen and Gao, Chang and Huang, Chengen and Lv, Chenxu and others},
  journal={arXiv preprint arXiv:2505.09388},
  year={2025}
}

@article{geor1,
  title={Geo-R1: Improving few-shot geospatial referring expression understanding with reinforcement fine-tuning},
  author={Zhang, Zilun and Guan, Zian and Zhao, Tiancheng and Shen, Haozhan and Cai, Yuxiang and Su, Zhonggen and Shang, Yongheng and Liu, Zhaojun and Yin, Jianwei and Li, Xiang},
  journal={ISPRS Journal of Photogrammetry and Remote Sensing},
  volume={237},
  pages={113--129},
  year={2026},
  publisher={Elsevier}
}

@article{sam3,
  title={Sam 3: Segment anything with concepts},
  author={Carion, Nicolas and Gustafson, Laura and Hu, Yuan-Ting and Debnath, Shoubhik and Hu, Ronghang and Suris, Didac and Ryali, Chaitanya and Alwala, Kalyan Vasudev and Khedr, Haitham and Huang, Andrew and others},
  journal={arXiv preprint arXiv:2511.16719},
  year={2025}
}
}

\end{document}